\documentclass[10pt,twocolumn,letterpaper]{article}

\usepackage{iccv}
\usepackage{times}
\usepackage{epsfig}
\usepackage{graphicx}
\usepackage{amsmath}
\usepackage{amssymb}
\usepackage{enumitem}
\usepackage{multirow}
\usepackage{booktabs}
\usepackage{graphicx}
\newcommand{\argmax}{\operatornamewithlimits{argmax}}
\newcommand{\argmin}{\operatornamewithlimits{argmin}}
\usepackage[T1]{fontenc}
\usepackage[dvipsnames]{xcolor}
\usepackage{authblk}

\usepackage[pagebackref=true,breaklinks=true,letterpaper=true,colorlinks,bookmarks=false]{hyperref}

\iccvfinalcopy 

\ificcvfinal\fi

\usepackage[capitalize]{cleveref}
\crefname{section}{Sec.}{Secs.}
\Crefname{section}{Section}{Sections}
\crefname{table}{Tab.}{Tabs.}
\Crefname{table}{Table}{Tables}

\usepackage{letltxmacro}
\LetLtxMacro\oldttfamily\ttfamily
\DeclareRobustCommand{\ttfamily}{\oldttfamily\csname ttsize\endcsname}
\newcommand{\setttsize}[1]{\def\ttsize{#1}}%

\begin{document}

\setttsize{\small}

\title{Mole Recruitment: Poisoning of Image Classifiers via Selective Batch Sampling}
\author{
Ethan Wisdom \quad Tejas Gokhale  \quad Chaowei Xiao \quad Yezhou Yang\\
Arizona State University\\ 
{\tt\small \{ewisdom, tgokhale, cxiao17, yz.yang\}@asu.edu}
}

\maketitle
\ificcvfinal\thispagestyle{empty}\fi

\begin{abstract}
In this work, we present a data poisoning attack that confounds machine learning models without any manipulation of the image or label.
This is achieved by simply leveraging the most confounding natural samples found within the training data itself, in a new form of a targeted attack coined ``\textit{Mole Recruitment}.''
We define moles as the training samples of a class that appear most similar to samples of another class, and show that simply re-structuring training batches with an optimal number of moles can lead to significant degradation in the performance of the targeted class.
We show the efficacy of this novel attack in an offline setting across several standard image classification datasets, and demonstrate the real-world viability of this attack in a continual learning (CL) setting.
Our analysis reveals that state-of-the-art models are susceptible to Mole Recruitment, thereby exposing a previously undetected vulnerability of image classifiers.
Code can be found here:
\url{http://github.com/wisdeth14/MoleRecruitment}

\end{abstract}

\section{Introduction}
\label{sec:introduction}
\begin{figure}
    \centering
    \includegraphics[width=\linewidth]{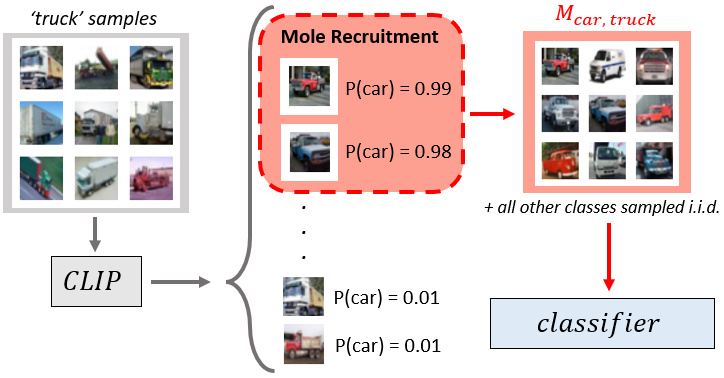}
    \caption{Using an external model such as CLIP, the distribution of the confounding class (truck) samples is examined with respect to their softmax probabilities of the attacked class (car). The samples with the highest probability are used to train an image classifier, which confounds the classifier's ability to distinguish between the attacked and confounding classes.}
    \label{fig:teaser}
\end{figure}
Continual learning, the ability to learn new tasks over time, is a pillar of human intelligence.
Machine learning models attempt to continually learn by training on new sets of data, pertaining to the new task, over discrete time steps.
However, training the model on new data may not only induce catastrophic forgetting, where the model forgets how to perform old tasks as it becomes biased to current data, but allows the potential for \textit{data poisoning} upon each new task, where data is maliciously presented to a model in order to shift its internal parameters so it fails upon test.
Since continual learning has been widely applied in safety-critical applications, including clinical diagnostics \cite{lee2020clinical} and autonomous driving \cite{cao2022autonomous, shaheen2022continual}, it is urgent to proactively understand these potential vulnerabilities.

In literature, traditional forms of poisoning leverage vast capabilities to impair a model's training via data manipulation (the value of a valid data sample is corrupted) and label flipping (a data sample becomes mislabeled) \cite{ramirez2022poisoning}.
Further stealthier methods involve designing and injecting seemingly innocuous, correctly labeled data that will skew the boundaries of a classifier to incur an accuracy penalty.
In the realm of image classification, this is commonly achieved through gradient-based methods, where additive pixel perturbations, imperceptible to the human eye, can induce poisonous effects \cite{shafahi2018poison, zhu2019transferable, munoz2017towards, yang2017generative, van2022poisoning}.

But what if an adversary could poison machine learning models using examples found within the training corpus itself?
Can a subset of the training data, without any additional pre-processing, manipulation, or optimization, be leveraged to confound a machine learning model?
In this work, we define \textit{``moles''} as training samples of one class that appear most similar to samples of another class.
We show that such moles within the dataset have the surprising ability to degrade model performance, and a selective batch sampling strategy that we call \textit{``Mole Recruitment''} can inflict large accuracy penalties of targeted classes.
We demonstrate that access to this batch sampling strategy is a significant threat to the robustness of image classifiers, and that an effective form of data poisoning can occur without any manipulation of the training samples.

In Mole Recruitment, given an image classifier trained on a particular set of data, moles are selected with the intention of confounding the model into believing data of a particular class holds the label of another.
Such poisonous data is selected by examining the samples of a particular class (the \textit{``confounding class''}) whose softmax probability is highest for that of a different class (the \textit{``attacked class''}).
No alterations of the data are required; we simply leverage existing samples from the original distribution.
By structuring batches with the optimal number of moles, to be appended to training data upon learning new tasks, we demonstrate we can exacerbate the effects of catastrophic forgetting by attacking classes previously learned.
This shows continual learning models are vulnerable to such attacks, further incentivizing the need for robustness in this field.

\noindent Our contributions and findings are organized as follows:
\begin{itemize}[nosep,noitemsep,leftmargin=*]
    \item In \cref{sec:method}, we present a new style of poisonous attack which leverages only valid, unaltered training data in order to degrade model accuracy of targeted classes.
    \item In \cref{sec:offline_exp}, to explore the feasibility of our attack, we operate in a controlled offline setting.
    We demonstrate its effectiveness upon deep neural networks, thus exposing new vulnerabilities of image classifiers.
    This attack produces up to a 50\% loss in per-class accuracy, and an optimized version of this attack produces up to a 15\% decrease in total model accuracy.
    \item In \cref{sec:continual} and \cref{sec:continual_exp}, we apply a model-agnostic form of this attack against state-of-the-art CL models, demonstrating a real-world threat model of Mole Recruitment. The attack achieves relative accuracy penalties of up to 48\% and 21\% of the attacked classes, for non-replay and replay methods, respectively.
\end{itemize}

\section{Related Work}
\label{sec:related_work}

\paragraph{Continual Learning.}

In continual learning, a model learns new, discrete tasks over time.
In image classification, this amounts to learning different sets of classes at different training steps.
Naive models are biased to perform best on most recently learned tasks, at the cost of overriding knowledge needed to perform on tasks previously learned, a phenomenon known as \textit{catastrophic forgetting}.
CL models attempt to mitigate catastrophic forgetting, although this can often inhibit the model's ability to adapt to new tasks, a trade-off known as \textit{stability vs. rigidity} \cite{de2021continual}.

\medskip\noindent\textbf{CL Methods.}
Some of the most empirically effective approaches in continual learning leverage experience replay by preserving a memory cache of exemplars from previous tasks \cite{prabhu2020gdumb, belouadah2019il2m, hou2019learning, wu2019large}.
By concurrently training with old samples, the model can avoid bias towards just the new task it learns.
Other works such as \cite{rebuffi2017icarl, yan2021dynamically}, leverage replay to maintain a feature extractor used to classify images.

Non-replay methods often utilize some form or parameter regularization, such as incorporating distillation loss \cite{li2017learning, dhar2019learning} or a penalty term to minimize the update of weights deemed crucial for previous tasks \cite{kirkpatrick2017overcoming, zenke2017continual}.
This mitigates forgetting by minimizing the $\ell_p$ norm shift from previous optima, but inhibits adaptability to learning new tasks (i.e. more stability, less plasticity).
More recent work uses a distillation scheme to reorganize an expansive feature representation, although requires a larger number of classes in the first task to ensure the initial representation is comprehensive \cite{zhu2022self}.
Other non-replay methodologies take a structure-based approach, and dynamically expand the network's architecture to incorporate new learning \cite{rusu2016progressive, schwarz2018progress}.

\medskip\noindent\textbf{Data Poisoning.}
Poisoning attacks aim to corrupt a model via the training data it is provided so that it fails upon inference.
Many effective poisoning methodologies take the form of clean-label attacks, and often use gradient-based approaches to create poisonous images through pixel-wise perturbations \cite{shafahi2018poison, zhu2019transferable, munoz2017towards, van2022poisoning}.
Other approaches use generative techniques, allowing an adversary to readily assemble a large quantity of poisonous images \cite{yang2017generative, munoz2019poisoning}.

\begin{figure*}
  \centering
  \includegraphics[width=0.49\linewidth]{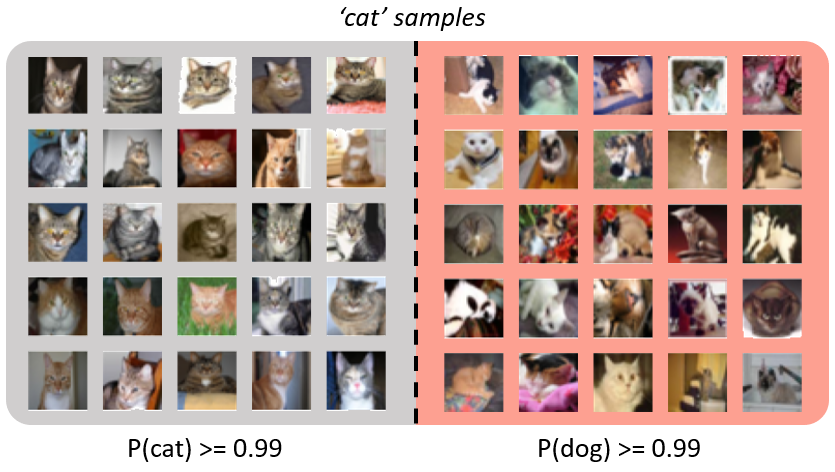}
  \includegraphics[width=0.49\linewidth]{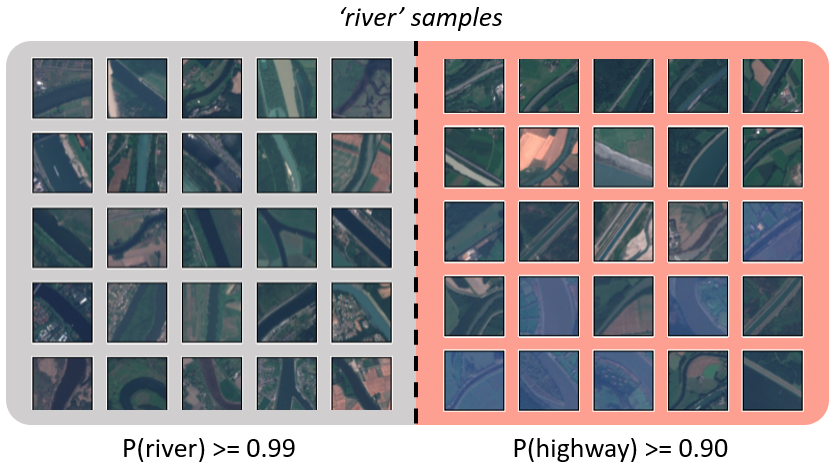}
  \caption{Visual example of top-25 moles (\textcolor{red}{red}) in contrast with samples the classifier accurately labels with high confidence (\textcolor{gray}{gray}). All images belong to the original training set and are not perturbed or transformed in any way. (Left) moles identified when $a{=}\texttt{dog}$ and $c{=}\texttt{cat}$; (Right) moles identified when $a{=}\texttt{highway}$ and $c{=}\texttt{river}$. Note that moles need not necessarily be visually confounding to humans, but do maximally confound the trained classifier.}
  \label{fig:viz_moles}
\end{figure*}

\section{Mole Recruitment}
\label{sec:method}

\subsection{Problem Setup and Notation.}

Consider an image classifier that is trained on a dataset $\mathcal{D}$ containing images $x_y$, where $y$ denotes the ground truth label of the image from label-set $\mathcal{Y}$ of size $k$.
The classifier learns a function $f$ that, given an image $x_y$, produces a set of softmax probabilities of the image's ground truth label for every label in $\mathcal{Y}$, i.e. $f(x_y)=\{P(y\!=\!0|x_y),\dots,P(y\!=\!k\!-\!1|x_y)\}$.
The classifier makes its prediction by selecting the label with the highest probability from the set.
Our main hypothesis is that images of one class can be used as a medium to confound the model's ability to classify samples from another class.
We designate the former as the \textit{``confounding class,''} or class $c$, and the latter as the \textit{``attacked class,''} or class $a$.
For any pair of classes \texttt{a,c} the most confounding image of class $c$ is the one for which the model outputs the largest probability for class $a$:
\begin{equation}
    m^\ast = \argmax_{x_c \in \mathcal{D}} P(y\!=\!a|x_c).
    \label{eq:mole}
\end{equation}
In our problem setting, the goal is to find such images, which we coin as \textit{``moles'',} and structure training batches in order to degrade the classifier's ability to distinguish between the attacked and confounding classes.
To effectively poison the classifier, we hypothesize numerous moles will be desired.
Let $\mathcal{M}^N_{a,c}$ be the set of top-$N$ moles that satisfy \cref{eq:mole} with confounding class $c$ and attacked class $a$; \cref{fig:viz_moles} shows a visual example of $\mathcal{M}^{25}_{a,c}$ for two standard image classification datasets.

\subsection{Selective Batch Sampling with Moles}
To preserve class-balance during training, 
our poisonous attack consists of the mole set (data from the confounding class) in addition to an equal number of randomly selected samples from every other class:
\begin{equation}
  \mathcal{D}_{poison} = \mathcal{M}^N_{a,c} \cup \{\mathcal{D}_i^N ~~ \forall i\in\mathcal{Y}  ~~ \mathrm{s.t.} ~ i\neq c\}.
  \label{eq:poison_set}
\end{equation}
The above set $\mathcal{D}_{poison}$ therefore represents a class-balanced poisonous training set, with the images from the confounding class being sampled non-i.i.d. by recruiting moles, i.e. images for which the classifier predicts high probabilities for class $a$, and images from all other $k{-}1$ classes sampled uniformly and i.i.d. from the training set.
This class-balancing is necessary to demonstrate any potential accuracy degradation is the result of the moles, and not gross class imbalance.

The essence of our attack strategy thus operates as follows:
Given a classifier trained on $\mathcal{D}$ with a pre-attack baseline accuracy $acc_{pre}$, we then derive the mole set $\mathcal{M}^N_{a,c}$ of size $N$ and construct the poisonous dataset $\mathcal{D}_{poison}$ according to \cref{eq:poison_set}.
The model is subsequently trained using $\mathcal{D}_{poison}$ in order to inflict a performance degradation of the attacked classes, resulting in post-attack accuracy $acc_{post}$.
We present an experiment on offline image classifiers in \cref{sec:case_study}, explore the relationship between the number, quality, and efficacy of the moles used (\cref{sec:tuning}), and leverage these findings to design an optimized version of our poisoning attack (\cref{sec:optimal}).

\begin{figure*}[t]
    \centering
    \includegraphics[width=0.33\linewidth]{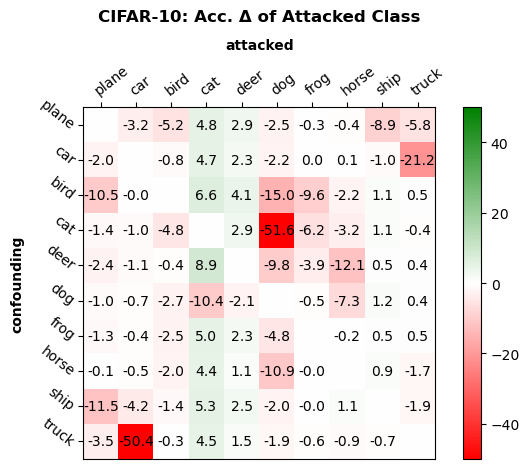}
    \includegraphics[width=0.33\linewidth]{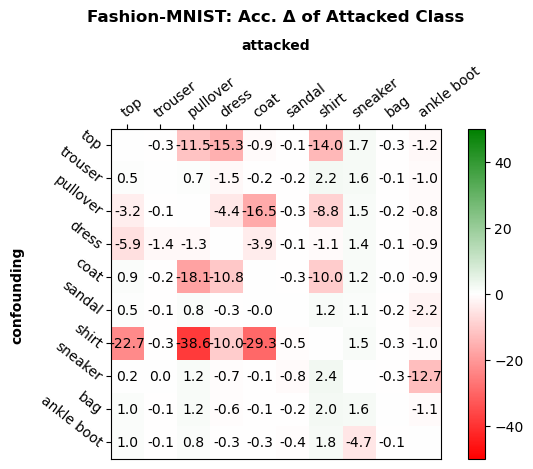}
    \includegraphics[width=0.33\linewidth]{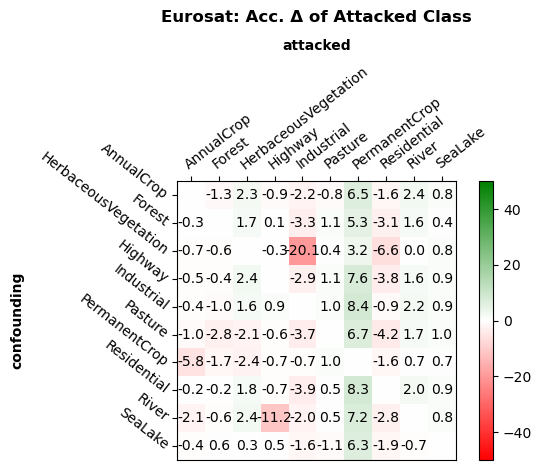}\\
    \includegraphics[width=0.33\linewidth]{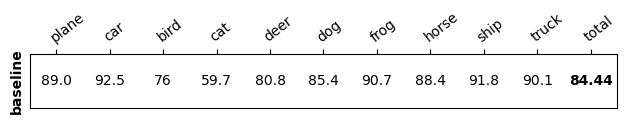}
    \includegraphics[width=0.33\linewidth]{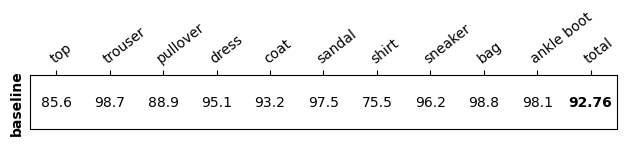}
    \includegraphics[width=0.33\linewidth]{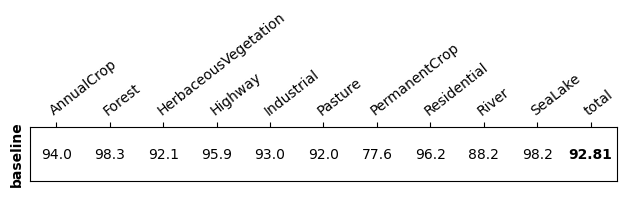}
    \caption{The change in accuracy of a trained ResNet-18 network after further training of the $\mathcal{M}^{100}_{a,c} \cup \{D_i^{100} ~~ \forall i\in\mathcal{Y}  ~~ \mathrm{s.t.} ~ i\neq c\}$ data for all possible combinations of $a$ and $c$ for the CIFAR-10, Fashion-MNIST, and EuroSAT datasets, averaged across 50 trials. Each value of the table reflects the change in accuracy of the attacked class for a particular \texttt{a,c} pair; e.g. when $a{=}\texttt{bird}$ and $c{=}\texttt{plane}$ the model's accuracy in predicting samples from the `bird' class drops by 5.2\%. Baseline accuracies prior to the poisonous attack are shown below. } 
    \label{fig:attacktables}
\end{figure*}

\section{Offline Experiments}
\label{sec:offline_exp}

\subsection{Poisoning of Offline Models} 
\label{sec:case_study}

\begin{table}
    \centering
    \resizebox{\linewidth}{!}{
    \begin{tabular}{@{}cr cccc@{}}
        \toprule
        \textbf{Model} & \textbf{Metric} & CIFAR-10 & Fashion-MNIST & EuroSAT\\
         \midrule
        \multirow{3}{*}{ResNet-18} 
         & $max(\Delta_{a})$ & $-55.1\pm{6.0}$ & $-38.6\pm{3.6}$ & $-21.1\pm{10.7}$\\
         & $95^{th}$ pctl.   & $-13.6\pm{1.5}$ & $-16.1\pm{1.0}$ & $-5.7\pm{1.2}$\\
         & $mean(\Delta_{a})$ &  $-2.7\pm{0.3}$ &  $-2.6\pm{0.1}$ & $-0.1\pm{0.2}$\\
         & $max(\Delta_{total})$ & $-5.2\pm{1.1}$ & $-3.6\pm{0.4}$ & $-7.0\pm{1.0}$\\
         \midrule
        \multirow{3}{*}{ResNet-50}
         & $max(\Delta_{a})$ & $-55.2\pm{4.5}$ & $-15.3\pm{1.5}$ & $-48.6\pm{22.9}$\\
         & $95^{th}$ pctl. & $-22.7\pm{2.0}$ &  $-7.7\pm{0.9}$ & $-6.4\pm{1.9}$\\
         & $mean(\Delta_{a})$ &  $-5.9\pm{0.3}$ &  $-0.9\pm{0.1}$ & $4.0\pm{0.4}$\\
         & $max(\Delta_{total})$ & $-5.9\pm{0.7}$ & $-0.7\pm{0.5}$ & $-9.7\pm{3.8}$\\
          \midrule
        \multirow{3}{*}{ViT}
         & $max(\Delta_{a})$ & $-42.8\pm{5.7}$ & $-54.9\pm{13.8}$ & $-31.7\pm{6.0}$ \\
         & $95^{th}$ pctl. & $-28.4\pm{3.1}$ & $-47.0\pm{18.0}$ & $-19.7\pm{3.6}$ \\
         & $mean(\Delta_{a})$ & $-12.8\pm{1.2}$ & $-16.8\pm{7.7}$ & $-5.1\pm{0.7}$ \\
         & $max(\Delta_{total})$ & $-12.2\pm{1.1}$ & $-16.2\pm{8.0}$ & $-6.3\pm{1.1}$ \\
        \bottomrule
    \end{tabular}
    }
    \vspace{1pt}
    \caption{Efficacy of offline poisoning attack on three benchmark datasets. Negative numbers imply drop in accuracy. \label{tab:benchmark_offline}}
\end{table}
Given a model trained on dataset $\mathcal{D}$, we obtain the mole sets $\mathcal{M}^N_{a,c}$ and dataset $D_{poison}$ for all combinations of attacked and confounding classes -- for a $k$-class classification problem there are $_kP_2$ possible permutations.
We then train the model additionally on each of these $\mathcal{D}_{poison}$ and compute the resulting accuracy degradation.
While 
offline models will optimally perform by jointly training on randomly shuffled training data, this experiment is a useful medium to explore the efficacy of Mole Recruitment before moving to a continual learning setting.

\medskip
\noindent\textbf{Experimental Setup.}
We use three benchmark datasets: Fashion-MNIST \cite{xiao2017fashion}, CIFAR-10 \cite{recht2018cifar}, and EuroSAT \cite{helber2019eurosat}, providing a variation in visual complexity to explore the effectiveness of the attack.
We use two convolution-based architectures, ResNet-18 and ResNet-50 \cite{he2016deep}, trained using SGD with batch-size 512, learning rate 0.01, momentum 0.9, and a learning rate scheduler, and the attention-based Vision Transformer (ViT) \cite{dosovitskiy2010image}, fine-tuned using batch-size 16 and learning rate 0.0002.
To demonstrate the impact of moles as a proof of concept, we arbitrarily select $N=100$, which amounts to only \textasciitilde0.2\% of all available training samples, and report results averaged over 50 trials.

\medskip\noindent\textbf{Evaluation Metrics.}
Let $\Delta_a$ be the change in the accuracy of the attacked class and $\Delta_{total}$ be the change in accuracy across all classes.
We then define the mean and maximum drop in accuracy of the attacked class over all \texttt{a,c} combinations as:
\begin{equation}
    mean(\Delta_a) = \underset{(a, c)}{\mathbb{E}}~\Delta_a, \quad 
    max(\Delta_a)  = \underset{(a, c)}{\mathrm{max}}~\Delta_a.
    \label{eq:metrics_attacked}
\end{equation}
We also report the 95$^{th}$ percentile of $\Delta_a$'s across all \texttt{a,c} pairs.
Lastly, we report the maximum total accuracy drop over all \texttt{a,c} pairs as:

\begin{equation}
    max(\Delta_{total}) = \underset{(a, c)}{\mathrm{max}}~\Delta_{total}.
    \label{eq:metrics_total}
\end{equation}

\paragraph{Results.}
The performance drop inflicted by our poisonous attack is shown in \cref{tab:benchmark_offline} on all three datasets.
For select \texttt{a,c} combinations, our simple selective sampling strategy incurs a large performance degradation for the attacked class, i.e. a $max(\Delta_a)$ of $-51.6\%$ for CIFAR-10 with ResNet-18.
As the model is selectively fed samples with high $P(y\!=\!a|x_c)$, samples whose features appear to belong to the attacked class despite holding the label of the confounding class, the internal parameters of the network experience shift. This shift is in a manner which confounds the model into believing subsequent samples from the attacked class belong to the confounding class, as evidenced by changes in the confusion matrix (see Appendix).
It should be noted most \texttt{a,c} combinations are ineffective, hence the low values of $mean(\Delta_{a})$ and $max(\Delta_{total})$. 
However, we argue severe degradation of selectively targeted classes is consequential, even if overall accuracy change is minimal.
We furthermore observe that performance degradation is larger for ResNet-50 than ResNet-18 for CIFAR-10 and EuroSAT, whereas it is smaller for Fashion-MNIST.
The poisonous attack also appears similarly effective against the attention-based ViT across all datasets, although its higher $mean(\Delta_{a})$ is inflated due to slight accuracy degradation across all classes (see Appendix), .

\cref{fig:attacktables} shows the heatmaps for attacked class accuracy penalties for all $_kP_2$ combinations in each dataset.
We observe \texttt{a,c} combinations that are more intuitively confusing i.e. (\texttt{dog, cat}) or (\texttt{car, truck}) deliver the highest accuracy penalties. 
Likewise, we observe that most upper-body garments (\texttt{top, shirt, coat, pullover, dress}) best confound each other.
On the other hand, combinations more obviously distinguishable to the human eye, i.e. (\texttt{frog, horse}), are ineffective.

\subsection{The Relationship between Mole Quality and Attack Efficacy}
\label{sec:tuning}
\begin{figure}[t]
  \centering
  \includegraphics[width=\linewidth]{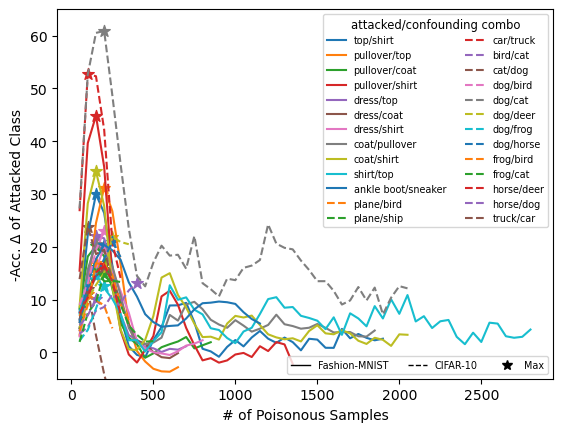}
  \caption{Change in accuracy of the attacked class vs. size $N$ of the mole set $\mathcal{M}^N_{a,c}$. 
}
   \label{fig:sizesweep}
\end{figure}

\begin{figure}[t]
  \centering
  \includegraphics[width=0.49\linewidth]{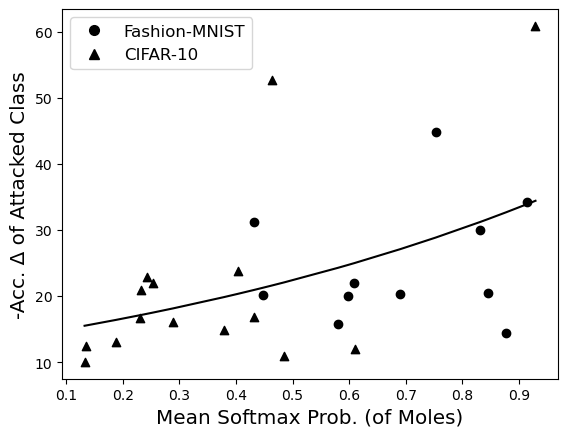}
  \includegraphics[width=0.49\linewidth]{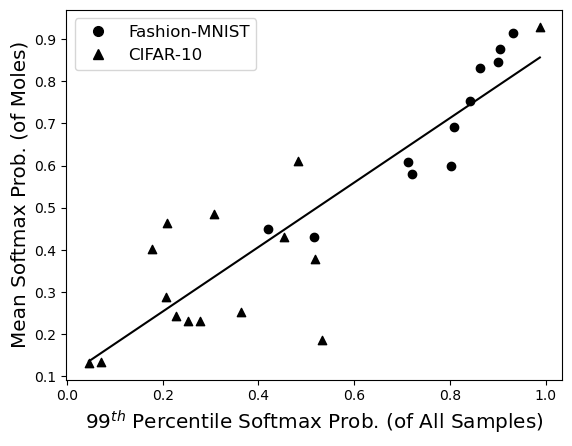}\\
  \caption{\textit{(Left)} Change in accuracy of the attacked class with respect to the mean $P(y\!=\!a|x_c)$ of the mole set. Data limited to the instances of $N$ at which the mole set delivers the maximum accuracy penalty for every \texttt{a,c} combination.
  \textit{(Right)} The mean $P(y\!=\!a|x_c)$ of the mole set (for the optimized number of moles for each \texttt{a,c} combination) vs. the 99$^{th}$ percentile of $P(y\!=\!a|x_c)$ probabilities for all confounding class samples.
  }
   \label{fig:distribution}
\end{figure}

We now explore how mole sets of different size $N$ and the distribution of the softmax probabilities $P(y\!=\!a|x_c)$, determine the effectiveness of the attack.
We begin by sweeping $N$ for each \texttt{a,c} pair, and record the change in accuracy of the attacked class as shown in \cref{fig:sizesweep}.
We observe that maximum accuracy penalties generally occur when the attack consists of only a few hundred samples from the confounding class, and the penalty decreases beyond this point.
As we increase $N$, the selective batch sampling strategy begins to select moles with lower $P(y\!=\!a|x_c)$, which have a lesser confounding effect.
This suggests the distribution of $P(y\!=\!a|x_c)$ used in the mole set influences the effectiveness of the poisonous attack.
Motivated to optimize the attack, we examine this relationship in \cref{fig:distribution} (left) by comparing the mean of the $P(y\!=\!a|x_c)$ distribution against the accuracy penalty of the attacked class, filtering our data to the instance of $N$ which delivers the maximum accuracy loss for each \texttt{a,c} combination.
We observe a loose correlation between these variables, where poisonous $P(y\!=\!a|x_c)$ distributions with higher means generally correlate to greater accuracy loss.

But what determines where a particular \texttt{a,c} class combination lies on the curve, i.e. why are some moles so much more effective at confounding the model than others?
This depends on the relationship between the attacked and confounding class, i.e. the distribution of $P(y\!=\!a|x_c)$ for all available samples of the confounding class.
Since we determine in \cref{fig:sizesweep} that an optimal attack results from only the few hundred most poisonous samples, yet there exists a large variation in the mean $P(y\!=\!a|x_c)$ of these samples as shown in \cref{fig:distribution} (left), we infer that the uppermost end of the $P(y\!=\!a|x_c)$ distribution of all confounding class samples bears influence on the samples needed to optimize the attack.
By comparing the 99$^{th}$ percentile of data samples from the confounding class, w.r.t.\ their attacked class softmax probability $P(y\!=\!a|x_c)$, against the mean $P(y\!=\!a|x_c)$ of the optimized attack for each \texttt{a,c} combination, we observe a very strong, linear correlation.
This correlation implies that to generate the optimal poisonous attack, the size $N$ of the mole set should be selected such that its mean $P(y\!=\!a|x_c)$ falls along the line in \cref{fig:distribution} (right).
While this correlation could likely vary with different datasets of different sizes, it demonstrates that a simple understanding of the softmax probabilities of natural data can be used to optimally confound a trained model.

\subsection{Optimized Mole Recruitment} 
\label{sec:optimal}

Here, we explore creating a mole set of size $N$ that can best confound a trained model based off the $P(y\!=\!a|x_c)$ distribution.
Furthermore, we explore compounding several mole sets, using multiple \texttt{a,c} combinations, into a single poisonous attack, to examine if total model accuracy can be further decreased from accuracy degradation of multiple classes.

Let $\rho_{a,c}$ be the 99$^{th}$ percentile of all $P(y\!=\!a|x_c)$ probabilities for a particular pair.
Let $\mu_{a,c}$ be the mean $P(y\!=\!a|x_c)$ of the mole set that delivers the maximum accuracy penalty of the attacked class.
We then derive $r(\cdot)$, the correlation between $\rho_{a,c}$ and $\mu_{a,c}$, using linear regression as shown in \cref{fig:distribution} (right).
We next derive the optimal mole set $\mathcal{M}^N_{a,c}$ , by selecting the top-$N$ moles such that:
\begin{equation}
  \small
  N = \argmin_{n} ~~ \frac{1}{n}\sum_{i=1}^{n} P(y\!=\!a| x^i_c) - r(\rho_{a,c}).
  \label{eq:optimized_set}
\end{equation}
We use multiple best-case mole sets from \cref{eq:optimized_set} to attack multiple classes in a single poisonous attack. Our method for determining these mole sets works as follows:

\begin{enumerate}[nosep,noitemsep]
  \item Rank the 99$^{th}$ percentile $\rho$ value of all possible \texttt{a,c} combinations.
  \item Select \texttt{a,c} combinations with highest $\rho$, avoiding any overlap of classes, until $\rho_{a,c} < \tilde{\rho}$, where $\tilde{\rho}$ acts as a threshold to omit poor quality combinations.
  \item Calculate the desired $\mu_{a,c}$ from $r(\rho_{a,c})$ and use \cref{eq:optimized_set} to determine the desired $N$ for each of the selected \texttt{a,c} pairs.
  \item Average each $N$ of the selected combinations and round to the nearest batchsize$/k$ to calculate $\tilde{N}$; this maintains the batchsize used for each gradient step of the attack.
 \end{enumerate}
The poisonous attack now takes the following form: 
\begin{equation}
    \small
    \mathcal{D}_{poison} = \{\mathcal{D}_i^{\tilde{N}} ~~ \forall i\in\mathcal{Y}  ~~ \mathrm{s.t.} ~ i\neq c_1, \dots, c_n\} \cup \bigcup_{i=1}^n \mathcal{M}^{\tilde{N}}_{a_i,c_i} .
  \label{eq:optimized_adv_set}
\end{equation}
We apply this methodology using $\tilde{\rho} = 0.1$ to formulate a poisonous attack against a trained model.
The results are shown in \cref{tab:compounding}.

\begin{table}
    \centering
    \resizebox{\linewidth}{!}{
    \begin{tabular}{|c|c|c|c|c|c|c|c|c|}
        \hline
        & \texttt{a,c} & $\rho_{a,c}$ & $N$ & $\Delta_a^{opt.}$ & $\Delta_a^{no}$ & $\Delta_{total}^{opt.}$ & $\Delta_{total}^{no}$\\
        \hline
        \multirow{4}{*}{\rotatebox{90}{\tiny\bf{CIFAR-10}}} 
        & {\scriptsize dog, cat} & 0.99 & 276 & -61.8 & -51.6 & \multirow{4}{*}{\bf{-14.0} } & -3.1\\
        & {\scriptsize deer, bird} & 0.52 & 135 & -4.6 & 4.1 & & -1.0\\
        & {\scriptsize car, truck} & 0.21 & 189 & -50.9 & -50.4 & & -5.1\\
        & {\scriptsize plane, ship} & 0.21 & 171 & -26.7 & -11.5 & & -0.3\\
        \hline
        \multirow{3}{*}{\rotatebox{90}{\tiny\bf{F-MNIST}}}  
        & {\scriptsize coat, shirt} & 0.93 & 282 & -19.9 & -29.3 & \multirow{3}{*}{ \bf{-4.8}} & -2.2\\
        & {\scriptsize ankle boot, sneaker} & 0.81 & 182 & -21.9 & -12.7 & & -0.9\\
        & {\scriptsize dress, top} & 0.71 & 137 & -17.5 & -15.3 & & -1.1\\
        \hline
    \end{tabular}
    }
    \vspace{1pt}
    \caption{Results (ResNet-18) of optimized Mole Recruitment. ($\Delta^{opt.})$ demonstrates its strength compared to the non-optimized ($\Delta^{no}$) version from \cref{fig:attacktables}.
    The combos are selected by the algorithm (in order of highest $\rho_{a,c}$) and their respective accuracy penalties are shown.
    $\tilde{N}=200$.
    \label{tab:compounding}
    }
\end{table}

We observe the attacked classes of each combination suffer accuracy degradation, demonstrating compounding separate mole sets into one attack is still effective.
Furthermore, we observe noticeably higher degradation in the model's overall accuracy than if a poisonous attack were to consist of only a single \texttt{a,c} pair.
While this methodology could be further optimized (different combo selection process to avoid suboptimal deer/bird combo, allow class imbalance for better $N$ selection, explore overlapping attacked and confounding used in other combinations), this experiment ultimately demonstrates a more effective poisonous attack.
By selecting combinations with higher upper-end $P(y\!=\!a|x_c)$ distributions, choosing the appropriate sample size based off the correlation study in \cref{sec:tuning}, and compounding these combinations into a single poisonous attack, we can achieve higher accuracy degradation compared to the `blind' approach in the \cref{sec:case_study}.

\section{Mole Recruitment in Continual Learning}
\label{sec:continual}
While the experiments in \cref{sec:method} yield drastic decreases in classifier accuracy of targeted classes, any model trained in an offline setting can avoid this attack by simply shuffling the training data.
Furthermore, the previous attacks were not model-agnostic, and required repeatedly querying the classifier to obtain the softmax probabilities in order to identify moles.
In this section, we demonstrate that continual learning is a real-world scenario in which Mole Recruitment poses a viable threat.
Additionally, by leveraging the neural network CLIP (Contrastive Language–Image Pre-training) \cite{radford2021learning} to obtain the softmax probabilities, we can implement Mole Recruitment such that no model access is required prior to the attack.

\begin{figure}
  \centering
  \includegraphics[width=\linewidth]{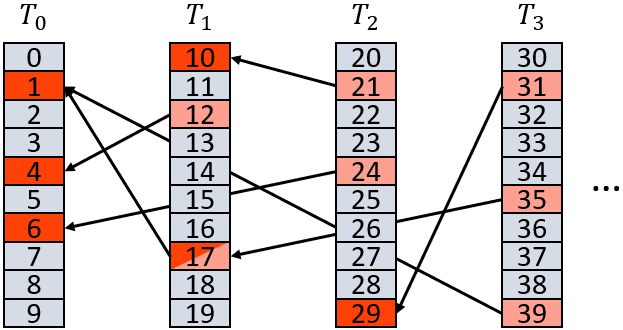}
  \caption{Visualization of our poisoning methodology in a continual learning setting. For all combinations where $r(\rho_{a,c}) > \tilde{\rho}$, a class from the current task (\textcolor{Salmon}{confounding}) will target a class learned in a previous task (\textcolor{red}{attacked}).}
  \label{fig:continual_visual}
\end{figure}

\medskip\noindent\textbf{Method.}
We consider a classifier that continually learns $N$ new tasks $T_0...T_{N-1}$,
where each task contains $k$ unique classes of images $x_y$, where $y$ denotes the ground truth label.
Thus, $[x_{ki},\dots,x_{k(i+1)-1}] \in T_i$ and the label-set $\mathcal{Y}$ is of size $kN$. 
The model learns in a \textit{class-incremental} fashion \cite{masana2022class}.
Thus, no task-ID is provided upon test, and the final classifier layer of the model is dynamically expanded to incorporate new neurons for new classes.

As visualized in \cref{fig:continual_visual}, we implement a Mole Recruitment attack under this CL setting, poisoning the classifier under the optimized conditions used in \cref{sec:optimal}.
For all classes in every new task learned, each acts as a potential confounding class $c$.
All classes from previously learned tasks are potential candidates to be the attacked class $a$.
Of these potential \texttt{a,c} pairs, all which have $r(\rho_{a,c}) > \tilde{\rho}$ are selected, the optimal $\tilde{N}$ is calculated, and the poison set $\mathcal{D}_{poison}$ is created.
Overlap of attacked classes is allowed, thus several confounding classes can be used as a medium to attack a particular attacked class.

To obtain the softmax probabilities $P(y\!=\!a|x_c)$ to determine $r(\rho_{a,c})$, however, we leverage CLIP.
Pre-trained on diverse image-caption pairings, CLIP is a neural network that can perform zero-shot image classification under any set of given text labels.
By providing CLIP the labels of our dataset, we have an alternative means of obtaining $P(y\!=\!a|x_c)$ that is agnostic to the victim classifier.
Upon creating $\mathcal{D}_{poison}$, minibatches from this poisonous set are appended to the training samples of the task, and are trained for a single epoch.

\medskip\noindent\textbf{Evaluation Metrics.}
We train a model for all tasks $T_1...T_N$ and compare the resulting final accuracies of all classes $y\in\mathcal{Y}$ between baseline performance ($acc_y^{b}$) and when the model is trained for an additional epoch of the poison set $\mathcal{D}_{poison}$ ($acc_y^{p}$) for all $T_i$ where $i > 1$.
To account for the low baseline performance of several models, we define $\Delta_{total}^{rel}$ as relative accuracy change of the entire model, and 
$\Delta_a^{rel}$ as the median of the set of relative accuracy changes of just the attacked classes in label-set $A$:
\begin{equation}
    \small
    \Delta_{total}^{rel} = mean(\{\frac{acc_y^{p} - acc_y^{b}}{acc_y^{b}} ~ \forall y \in \mathcal{Y}\})
    \label{eq:metrics_rel_total}
\end{equation}
\begin{equation}
    \small
    \Delta_{a}^{rel} = median(\{\frac{acc_y^{p} - acc_y^{b}}{acc_y^{b}} ~ \forall y \in A\})
    \label{eq:metrics_rel_attacked}
\end{equation}
The quantity of attacked classes (the size of set $A$) and the quantity of confounding classes are denoted as \#$_a$ and \#$_c$, respectively.

\begin{table}    
    \centering
    \small\addtolength{\tabcolsep}{-2.5pt}
    \begin{tabular}{|c|cc|ccll|}
    \hline
    \multicolumn{7}{|c|}{\textbf{CIFAR-100}} \\
    \hline
    \multicolumn{1}{|c}{} & \multicolumn{1}{c}{method} & \multicolumn{1}{c|}{$\tilde{\rho}$} & \multicolumn{1}{c}{$\#_a$} & \multicolumn{1}{c}{$\#_c$} & \multicolumn{1}{c}{$\Delta_{a}^{rel}$} & \multicolumn{1}{c|}{$\Delta_{total}^{rel}$} \\
    \hline
    \multirow{9}{*}{\rotatebox{90}{Non-Replay}} & \multirow{3}{*}{LwF \cite{li2017learning}} & 0.5 & $18$ & $32$ & $\mathbf{-17.9}\pm{21.6}$ & $-8.5\pm{3.5}$ \\
    & & 0.1 & $31$ & $60$ & $-4.8\pm{8.5}$ & $-2.1\pm{5.3}$ \\
    & & 0.01 & $32$ & $67$ & $-10.1\pm{7.6}$ & $-4.8\pm{4.8}$ \\
    \cline{2-7}
    & \multirow{3}{*}{LwM \cite{dhar2019learning}} & 0.5 & $6$ & $7$ & $-42.5\pm{8.2}$ & $-2.2\pm{4.8}$ \\
    & & 0.1 & $11$ & $16$ & $-39.6\pm{21.5}$ & $-5.8\pm{3.8}$ \\
    & & 0.01 & $11$ & $17$ & $\mathbf{-48.6}\pm{13.2}$ & $-4.5\pm{2.6}$ \\
    \cline{2-7}
    & \multirow{3}{*}{SSRE\footnotemark[1] \cite{zhu2022self}} & 0.5 & 21 & 29 & $-6.6\pm{1.0}$ & $-18.2\pm{1.1}$\\
    & & 0.1 & 34 & 49 & $-6.9\pm{1.8}$ & $-18.8\pm{1.3}$ \\
    & & 0.01 & 34 & 50 & $\mathbf{-7.3}\pm{2.2}$ & $-18.9\pm{1.2}$ \\
    \hline
    \multirow{9}{*}{\rotatebox{90}{Replay}} & \multirow{3}{*}{iCaRL \cite{rebuffi2017icarl}} & 0.5 & $24$ & $39$ & $-8.7\pm{4.3}$ & $-1.1\pm{2.0}$ \\
    & & 0.1 & $40$ & $76$ & $\mathbf{-12.2}\pm{4.7}$ & $-10.3\pm{5.6}$ \\
    & & 0.01 & $42$ & $85$ & $-6.4\pm{3.8}$ & $-3.9\pm{3.0}$ \\
    \cline{2-7}
    & \multirow{3}{*}{IL2M \cite{belouadah2019il2m}} & 0.5 & $24$ & $39$ & $-6.6\pm{17.4}$ & $-6.1\pm{9.9}$ \\
    & & 0.1 & $40$ & $75$ & $\mathbf{-11.8}\pm{15.2}$ & $-9.7\pm{12.9}$ \\
    & & 0.01 & $42$ & $84$ & $-11.1\pm{11.6}$ & $-11.0\pm{5.8}$ \\
    \cline{2-7}
    & \multirow{3}{*}{DER \cite{yan2021dynamically}} & 0.5 & 27 & 40 & $\mathbf{-12.8}\pm{3.0}$ & $-4.4\pm{1.7}$ \\
    & & 0.1 & 43 & 78 & $-12.4\pm{2.6}$ & $-4.0\pm{0.9}$ \\
    & & 0.01 & 45 & 89 & $-12.2\pm{2.5}$ & $-4.0\pm{0.9}$ \\
    \hline
    \end{tabular}
    \vspace{1pt}
    \caption{The change in final accuracy due to Mole Recruitment. Accuracy is reported after all tasks are learned, averaged across 5 trials. Classes with final baseline accuracies less than 10\% are filtered out to reduce noise. Baseline performance for CIFAR-100: {LwF: $29.3\pm{0.9}$, LwM: $21.2\pm{1.3}$, SSRE: $60.6\pm{0.3}$, iCaRL: $35.0\pm{2.1}$, IL2M: $40.5\pm{2.3}$, DER: $67.0\pm{0.4}$}.}
    \label{tab:continual_table_cifar}
\end{table}

\paragraph{Continual Learning Models}
\label{sec:cl_models}

We examine the impact of the poisonous attack on both traditional and recent state-of-the-art CL models.
We also use a mix of exemplar-free models, along with models that leverage experience replay.
All models are used to learn in a class-incremental manner, and borrow the framework used in \cite{masana2022class} (LwF, LwM, iCaRL, IL2M) and \cite{zhou2021pycil} (SSRE, DER).

\section{Continual Learning Experiments}
\label{sec:continual_exp}

\paragraph{Datasets.}
We use two benchmark datasets, standard to class-incremental CL literature: CIFAR-100 \cite{recht2018cifar} and Imagenet-Subset, the latter of which contains the first 100 classes of the ImageNet dataset \cite{russakovsky2015imagenet}.
Each dataset is randomly split into 10 tasks, with 10 classes per task.\footnote{\label{myfootnote}SSRE's first task is composed of 50 classes to establish its feature representation; the remaining 5 tasks each consist of 10 classes.}
For CIFAR-100, the CL models are trained with ResNet-32 architecture, batch-size 128, and 200 epochs per task.
For ImageNet-Subset, the CL models are trained with ResNet-18 architecture, batch-size 256, and 100 epochs per task.

\paragraph{Training setup.}
In accordance with \cite{masana2022class}, the Continual Hyperparameter Framework grid search is used to determine the learning rate on a per-task basis.
The models are trained with SGD, momentum of 0.9 and weight decay of 0.0002.
The model trains off the poisonous set $\mathcal{D}_{poison}$ for a single epoch, and for methodologies that use replay, training is done in conjunction with the saved exemplars.
Exemplar memory is fixed to 2000 images throughout training, thus the exemplars per class decrease as more tasks are learned.

\begin{table}    
    \centering
    \small\addtolength{\tabcolsep}{-2.5pt}
    \begin{tabular}{|c|cc|ccll|}
    \hline
    \multicolumn{7}{|c|}{\textbf{ImageNet-Subset}} \\
    \hline
    \multicolumn{1}{|c}{} & \multicolumn{1}{c}{method} & \multicolumn{1}{c|}{$\tilde{\rho}$} & \multicolumn{1}{c}{$\#_a$} & \multicolumn{1}{c}{$\#_c$} & \multicolumn{1}{c}{$\Delta_{a}^{rel}$} & \multicolumn{1}{c|}{$\Delta_{total}^{rel}$} \\
    \hline
    \multirow{9}{*}{\rotatebox{90}{Non-Replay}} & \multirow{3}{*}{LwF \cite{li2017learning}} & 0.5 & $29$ & $42$ & $-11.3\pm{3.7}$ & $-11.6\pm{3.3}$ \\
    & & 0.1 & $35$ & $59$ & $\mathbf{-18.4}\pm{7.7}$ & $-13.5\pm{3.1}$ \\
    & & 0.01 & $37$ & $66$ & $-8.8\pm{9.0}$ & $-9.9\pm{2.6}$ \\
    \cline{2-7}
    & \multirow{3}{*}{LwM \cite{dhar2019learning}} & 0.5 & $14$ & $18$ & $-18.6\pm{24.7}$ & $~~~2.9\pm{9.9}$ \\
    & & 0.1 & $18$ & $26$ & $-8.2\pm{10.6}$ & $~~~1.3\pm{8.2}$ \\
    & & 0.01 & $18$ & $28$ & $\mathbf{-22.7}\pm{15.3}$ & $-1.5\pm{6.9}$ \\
    \cline{2-7}
    & \multirow{3}{*}{SSRE\footnotemark[1] \cite{zhu2022self}} & 0.5 & $32$ & $39$ & $-3.5\pm{0.6}$ & $-17.7\pm{1.0}$ \\
    & & 0.1 & $38$ & $50$ & $-2.5\pm{0.2}$ & $-17.3\pm{0.8}$ \\
    & & 0.01 & $38$ & $50$ & $\mathbf{-3.6}\pm{1.0}$ & $-18.0\pm{0.9}$ \\
    \hline
    \multirow{9}{*}{\rotatebox{90}{Replay}} & \multirow{3}{*}{iCaRL \cite{rebuffi2017icarl}} & 0.5 & $38$ & $57$ & $-15.0\pm{3.2}$ & $-10.0\pm{3.6}$ \\
    & & 0.1 & $47$ & $79$ & $-18.3\pm{4.7}$ & $-14.5\pm{3.8}$ \\
    & & 0.01 & $46$ & $85$ & $\mathbf{-18.6}\pm{3.8}$ & $-14.5\pm{2.4}$ \\
    \cline{2-7}
    & \multirow{3}{*}{IL2M \cite{belouadah2019il2m}} & 0.5 & $30$ & $41$ & $\mathbf{-4.1}\pm{13.3}$ & $-2.0\pm{5.0}$ \\
    & & 0.1 & $38$ & $61$ & $-3.4\pm{10.9}$ & $-4.9\pm{3.0}$ \\
    & & 0.01 & $40$ & $61$ & $~~~~0.7\pm{8.8}$ & $-8.2\pm{2.0}$ \\
    \cline{2-7}
    & \multirow{3}{*}{DER \cite{yan2021dynamically}} & 0.5 & $42$ & $56$ & $\mathbf{-21.0}\pm{1.0}$ & $-10.2\pm{1.1}$ \\
    & & 0.1 & $49$ & $81$ & $-20.5\pm{1.3}$ & $-10.3\pm{1.1}$ \\
    & & 0.01 & $50$ & $89$ & $-20.4\pm{1.2}$ & $-10.2\pm{1.1}$ \\
    \hline
    \end{tabular}
    \vspace{1pt}
    \caption{Baseline performance for ImageNet-Subset: {LwF: $40.0\pm{2.7}$, LwM: $28.5\pm{3.6}$, SSRE: $66.6\pm{0.3}$, iCaRL: $44.7\pm{2.7}$, IL2M: $39.0\pm{0.7}$, DER: $72.0\pm{0.5}$}.}
    \label{tab:continual_table_imagenet}
\end{table}

\paragraph{Results.}
The impact of using Mole Recruitment to attack continual learning models, relative to baseline performance, is recorded in \cref{tab:continual_table_cifar} and \cref{tab:continual_table_imagenet}.
While the impact to total classifier accuracy is minimal, we observe Mole Recruitment is largely effective against targeted classes, achieving up to 48.6\% and 22.7\% $\Delta_{a}^{rel}$ relative attacked accuracy loss for non-replay methods, for CIFAR-100 and Imagnet-Subset, respectively.
Methods that have the advantage of memory storage are generally more robust, and moles only incur a $\Delta_{a}^{rel}$ loss of up to 12.2\% and 21.0\% for CIFAR-100 and Imagnet-Subset, respectively.

We also observe that $\Delta_{a}^{rel}$ is highly variant for non-replay methods, and less so for replay methods.
We hypothesize that replay methods learn in a much more controlled setting as the network is concurrently trained with classes already seen, which reduces internal parameter shift.
In contrast, we hypothesize that the parameters of non-replay methods, which have lower baseline performance, are significantly more subject to noise from learning data of new classes.
SSRE is a significant exception to this case; we hypothesize that by initializing training with 5 times the number of classes for its first task, it establishes a feature representation more robust to poisoning compared to the other methods that learn in a true incremental fashion.
Curiously, however, SSRE suffers greater total accuracy degradation from the attack.

As expected, we also observe that as the value of $\tilde{\rho}$ is lowered, more classes are attacked as more \texttt{a,c} pairs with weaker $\rho_{a,c}$ are selected.
The impact of $\tilde{\rho}$ on accuracy degradation varies.
On non-replay methods, the combinations selected with $\tilde{\rho}=0.5$ seem sufficient to deliver high accuracy degradation.
For some non-replay methods, which are more robust, using a lower $\tilde{\rho}$ seems to be preferred to attack more classes for potential degradation, even though the $\rho_{a,c}$ value of these combinations are weaker.

Lastly, we reiterate that the moles in this attack, unlike in \cref{sec:method}, were identified via CLIP.
The results in \cref{tab:continual_table_cifar} and \cref{tab:continual_table_imagenet} show that by leveraging a generalized image-recognition network, we can still generate effective attacks in a model-agnostic manner.

\section{Analysis and Reflection}
\label{sec:analysis}

\paragraph{Further Observations of the Offline Experiments.}
While effective \texttt{a,c} combinations are easily identifiable from the the heatmap in \cref{fig:attacktables}, we observe that the inverse of certain combinations are much less effective.
For eg., in CIFAR-10, (\texttt{car, truck}) induces a 50.4\% drop in car accuracy while (\texttt{truck, car}) only induces in a 21.2\% drop in truck accuracy.
This implies car-like trucks are weaker moles than truck-like cars.
We also observe effective combinations that are less intuitively confusing, i.e. (\texttt{plane, bird}).
We attribute this to the spurious correlation introduced by the sky in the background of these images.
We also observe that attacking \texttt{cat} leads to an improvement in the model's \texttt{cat} accuracy, likely a result of the model's low baseline \texttt{cat} accuracy of 60\%.

We note there is a negligible change in accuracy of the non-attacked classes (see Appendix).
However, the confounding class generally experiences some accuracy \textit{improvement} for all \texttt{a,c} combinations.
We postulate that biasing the training with low-confidence samples of the confounding class (samples with low $P(y\!=\!c|x_c)$) improves robustness to edge cases.

Furthermore, we observe total accuracy change of the model after the attack is usually negligible, even after large shifts in performance of the attacked class, since only one of the $k$ classes is degraded (and the confounding class can experience some improvement).
This implies attacks leading to imbalances in accuracy could go unnoticed if users only give attention to the overall accuracy of the model.

\paragraph{Moles in a Continual Learning Environment.}
The results of of \cref{sec:continual} demonstrate the threat of Mole Recruitment in the continual learning domain.
Exemplar-free CL models, which demonstrated the greatest vulnerability, must be particularly wary of such poisoning attacks, and understand the risk of accepting naturally existing images for training which the machine may confound for a class it has already learned.
This is especially true in a \textit{online} continual learning environment, where training is less structured as the model learns off a continuous stream of data belonging to tasks both old or new.
If an adversary were to feed a mole to such a model at any point in time, its current state would be biased to perform poorly for the attacked class.

While CL models that utilize replay demonstrate a degree of robustness to this poisoning, these approaches may lack feasibility in many real-world settings.
The need to store samples of an indefinite number of new tasks challenges scalability, and in some use cases storage may not be allowed in the event of data privacy concerns.

\section{Conclusion}
\label{sec:conclusion}

In this study, we present a new style of poisonous attack by recruiting moles -- data samples of one class that have the appearance of samples from another -- that exist within natural, training data.
Our results reveal just how sensitive neural networks can behave in response to their training inputs, even when valid data samples are used: in the offline experiments, a mere two gradient steps could deliver accuracy degradation of over 50\% for the targeted class.
We then refined this methodology by compounding class combinations with the optimal number of mole samples and demonstrated how continual learning models are susceptible to degradation from this attack, revealing a real-world threat to CL computer vision models.

This work also opens new avenues to explore the dynamics behind Mole Recruitment, for instance, understanding why certain class combinations make for effective moles through feature-level explanations and their connection to spurious biases in the dataset.
Furthermore, methods to mitigate and prevent the adverse effects of Mole Recruitment need to be explored.
The most obvious response would be to establish a filter to prevent moles from being included within the training set.
However, these edge case samples can also be helpful in providing robustness to a model's performance upon inference, and thus in certain instances, paradoxically, moles could instead be desired.
Thus, a clear dilemma may exist when choosing whether to accept or reject training data, even if the image is deemed valid and unaltered.
These decision boundaries will need to be further understood in order to properly defend against poisonous attacks in a continual learning domain.

\medskip
\noindent {\bf Acknowledgements:} This work was supported by NSF grants  \#1750082, \#2101052. EW was partially supported by ASU Fulton Engineering Schools' Dean's fellowship.

\newpage
{\small
\bibliographystyle{ieee_fullname}
\bibliography{old, new, tgokhale}
}

\clearpage
\appendix
\section*{Appendix}
\section{Confusion Matrices}

\begin{figure*}[h]
    \centering
    \includegraphics[width=0.49\linewidth]{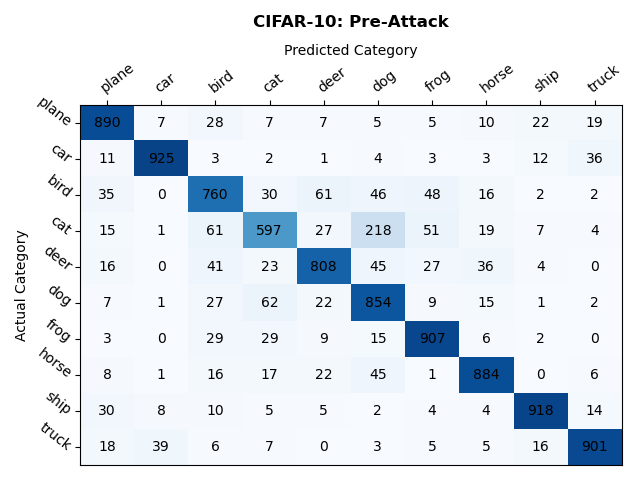}
    \includegraphics[width=0.49\linewidth]{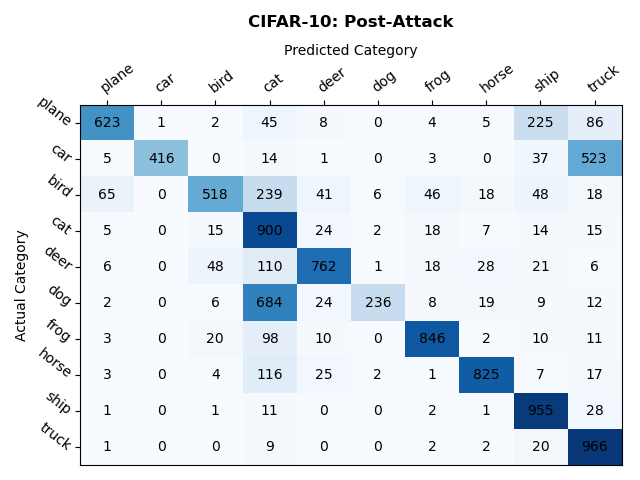}\\
    \includegraphics[width=0.49\linewidth]{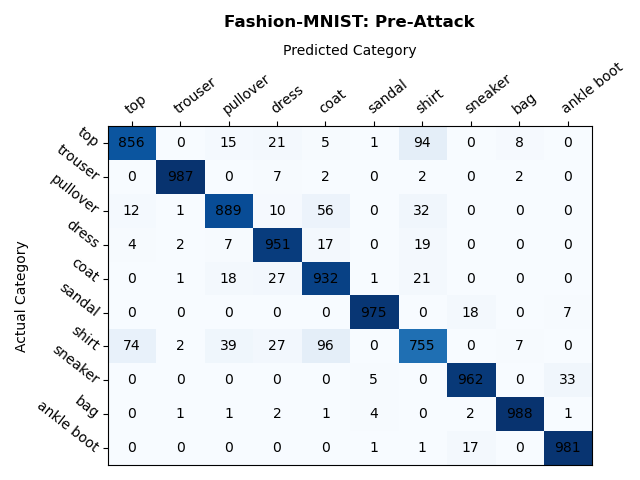}
    \includegraphics[width=0.49\linewidth]{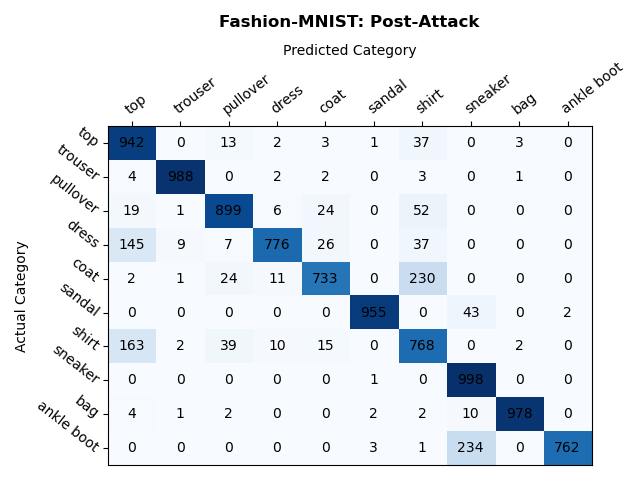}\\
    \caption{The confusion matrices, for CIFAR-10 and Fashion-MNIST, before and after the optimized attack as described in Sec. 4.3, where multiple attacked/confounding class/combinations are compounded into a single attack. For CIFAR-10, the moles consist of $a{=}\texttt{dog}$, $c{=}\texttt{cat}$; $a{=}\texttt{deer}$, $c{=}\texttt{bird}$; $a{=}\texttt{car}$, $c{=}\texttt{truck}$; $a{=}\texttt{plane}$, $c{=}\texttt{ship}$. For Fashion-MNIST, the moles consist of $a{=}\texttt{coat}$, $c{=}\texttt{shirt}$; $a{=}\texttt{ankle boot}$, $c{=}\texttt{sneaker}$; $a{=}\texttt{dress}$, $c{=}\texttt{top}$. We observe the attack generally works as intended as, upon inference, the model more often classifies samples from the attacked class as belonging to the confounding class. For example, in CIFAR-10, when $a{=}\texttt{dog}$, $c{=}\texttt{cat}$, over 600 more \texttt{dog} samples are misclassified as \texttt{cat}.}
    \label{fig:confusion}
\end{figure*}

In \cref{fig:confusion}, we provide confusion matrices that demonstrate how the model misclassifies data samples due to the mole recruitment attack.
The confusion matrices represent the optimized attack described in Sec. 4.3, where multiple attacked/confounding class/combinations are compounded into a single attack.
We observe, after the attack, that samples from the attacked class are frequently misclassified as belonging to the confounding class.
This demonstrates our attack strategy works as intended: confound the model in it's ability to distinguish between two particular classes by selectively sampling samples from the confounding class that, from a softmax probability perspective, appear to belong to the attacked class.

\section{Heatmaps}

\begin{figure*}[t]
    \centering
    \includegraphics[width=0.30\linewidth]{figures/tablecifarattacked.png}
    \includegraphics[width=0.30\linewidth]{figures/tablefashionattacked.png}
    \includegraphics[width=0.30\linewidth]{figures/tableeurosatattacked.png}\\
    \includegraphics[width=0.30\linewidth]{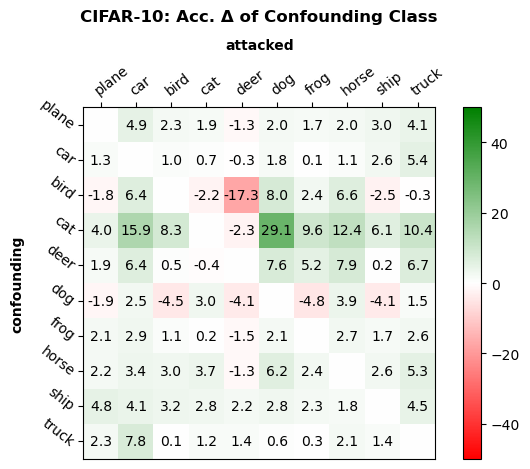}
    \includegraphics[width=0.30\linewidth]{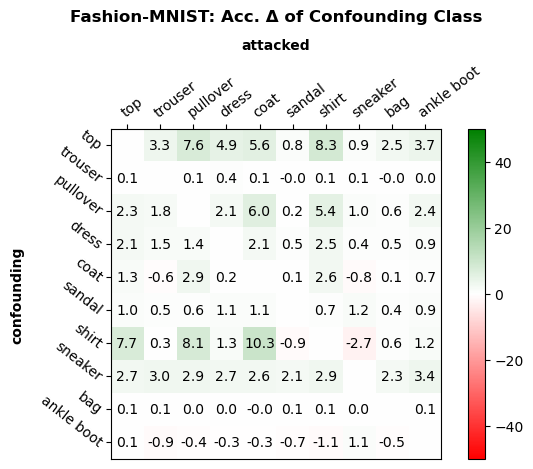}
    \includegraphics[width=0.30\linewidth]{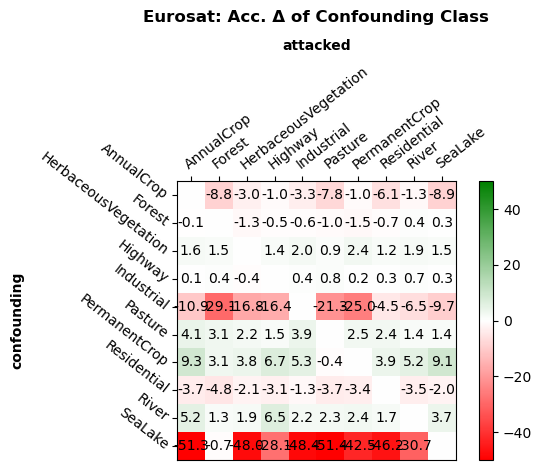}\\
    \includegraphics[width=0.30\linewidth]{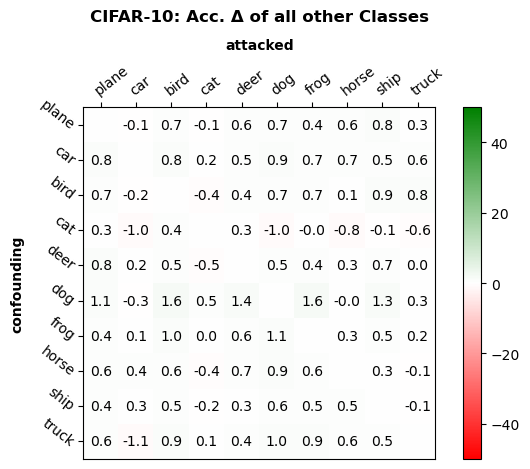}
    \includegraphics[width=0.30\linewidth]{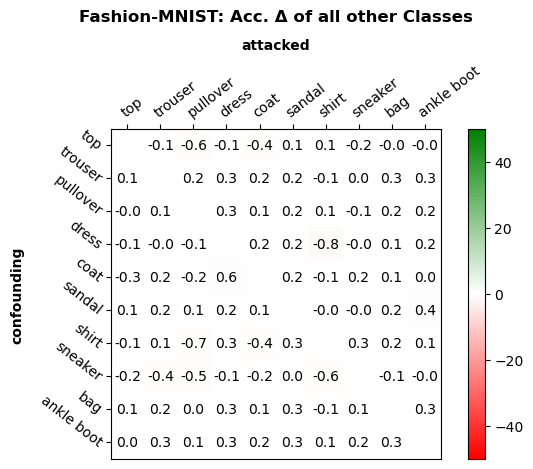}
    \includegraphics[width=0.30\linewidth]{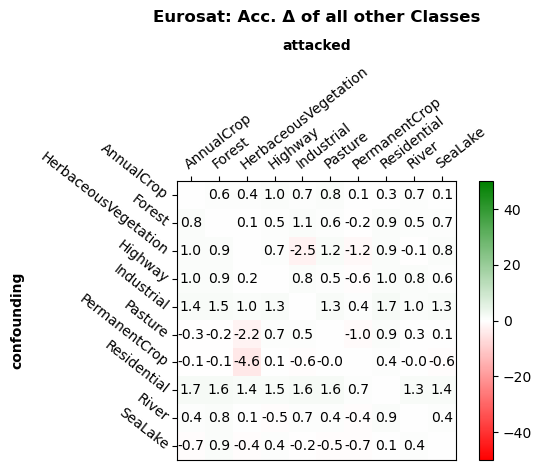}\\
    \includegraphics[width=0.30\linewidth]{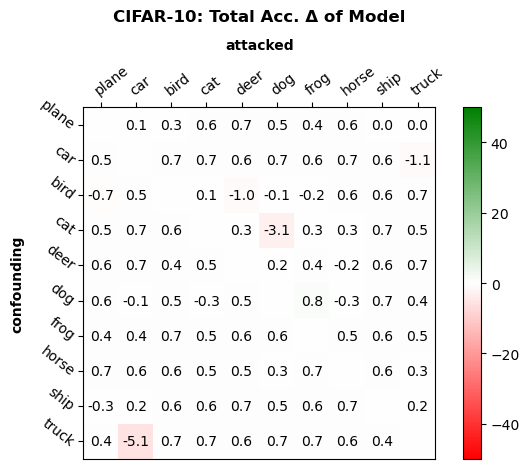}
    \includegraphics[width=0.30\linewidth]{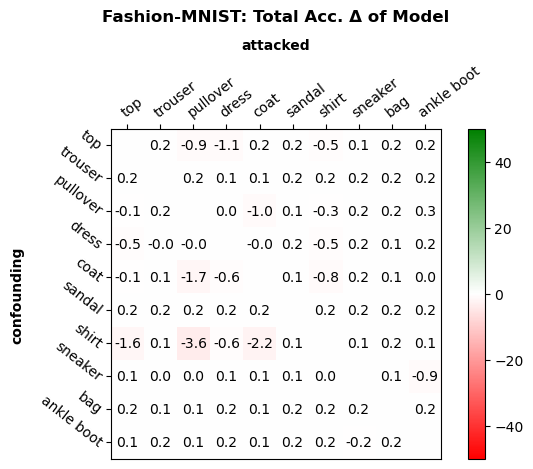}
    \includegraphics[width=0.30\linewidth]{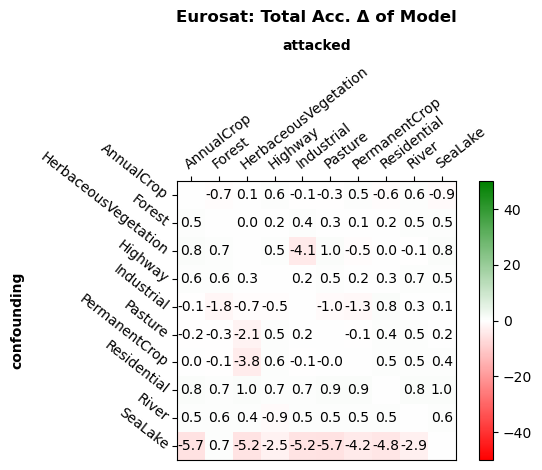}\\
    \includegraphics[width=0.33\linewidth]{figures/baselinecifar.png}
    \includegraphics[width=0.33\linewidth]{figures/baselinefashion.png}
    \includegraphics[width=0.33\linewidth]{figures/baselineeurosat.png}
    \caption{Heatmaps for all classes from the offline attack in Sec. 4.1 for ResNet-18 architecture. Ordered from left to right: CIFAR-10, Fashion-MNIST, and EuroSAT datasets. Ordered from top to bottom: accuracy change of the attacked class, the confounding class, the average of the eight non-attacked/non-confounding classes, and total accuracy of the classifier. Baseline accuracies of the model prior to the poisonous attack are shown below.}
    \label{fig:heatmaps18}
\end{figure*}

\begin{figure*}[t]
    \centering
    \includegraphics[width=0.30\linewidth]{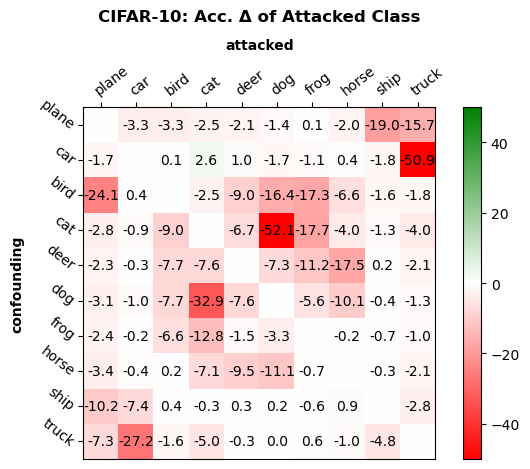}
    \includegraphics[width=0.30\linewidth]{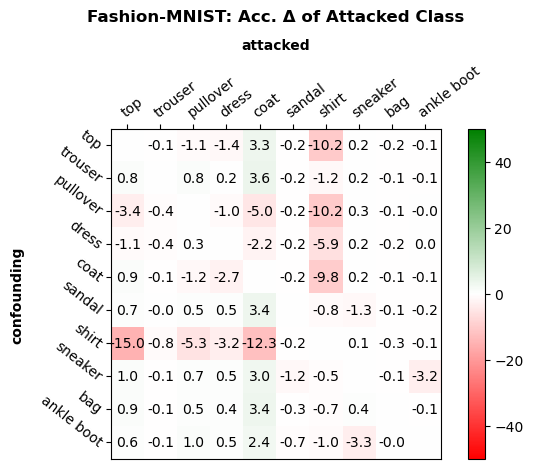}
    \includegraphics[width=0.30\linewidth]{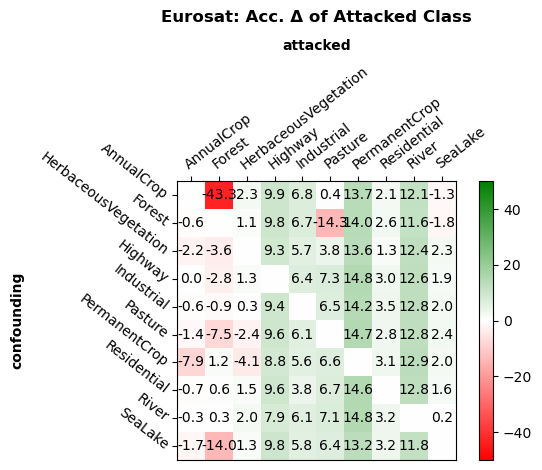}\\
    \includegraphics[width=0.30\linewidth]{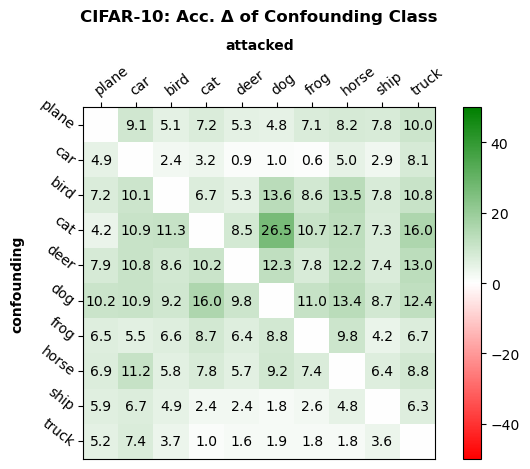}
    \includegraphics[width=0.30\linewidth]{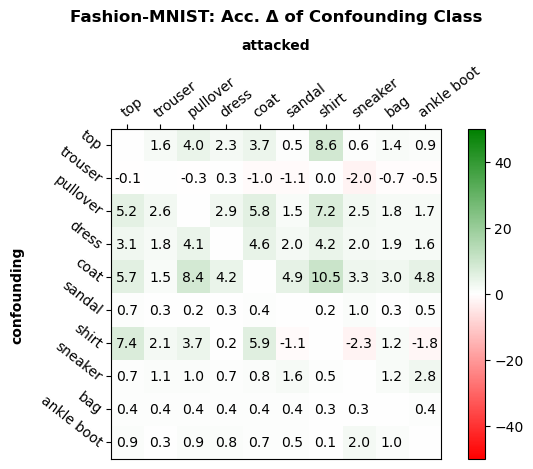}
    \includegraphics[width=0.30\linewidth]{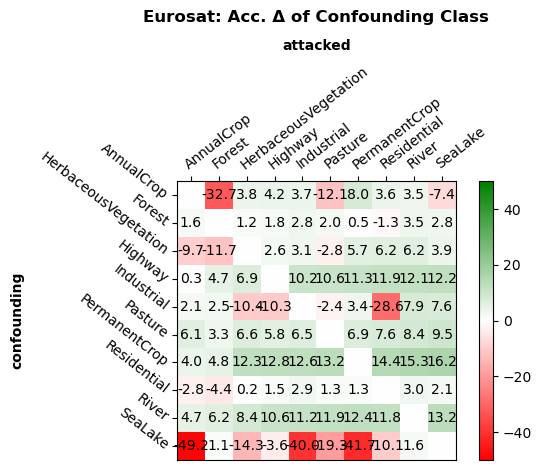}\\
    \includegraphics[width=0.30\linewidth]{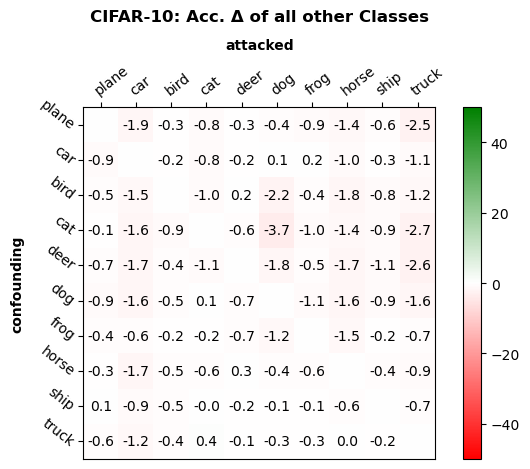}
    \includegraphics[width=0.30\linewidth]{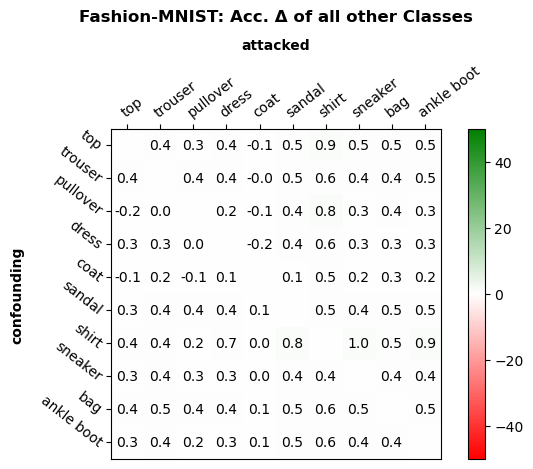}
    \includegraphics[width=0.30\linewidth]{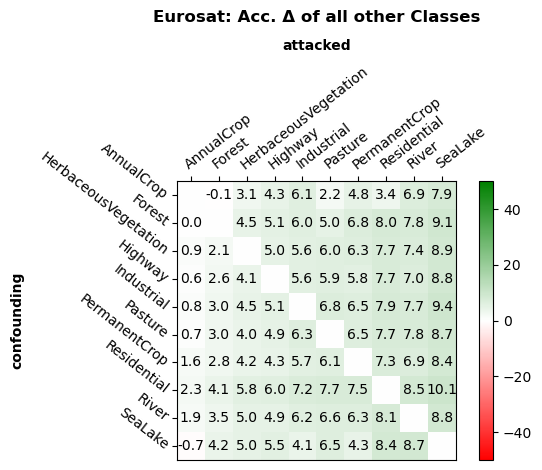}\\
    \includegraphics[width=0.30\linewidth]{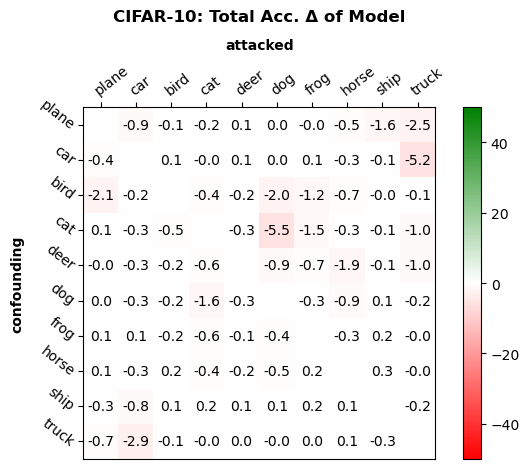}
    \includegraphics[width=0.30\linewidth]{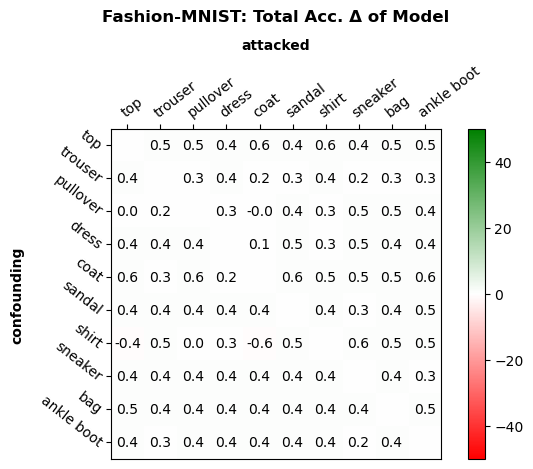}
    \includegraphics[width=0.30\linewidth]{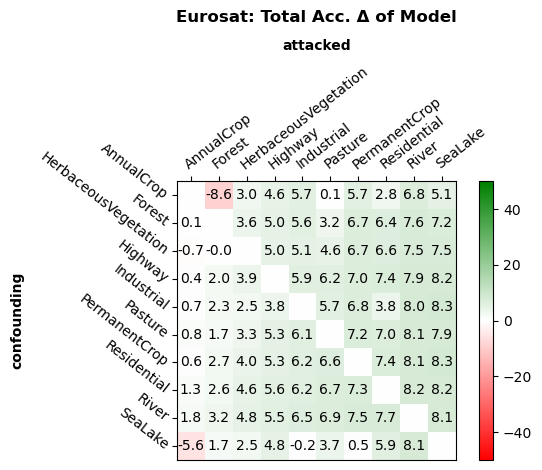}\\
    \includegraphics[width=0.33\linewidth]{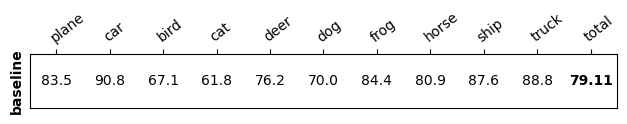}
    \includegraphics[width=0.33\linewidth]{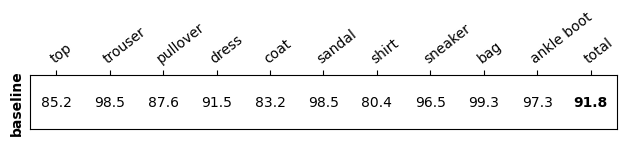}
    \includegraphics[width=0.33\linewidth]{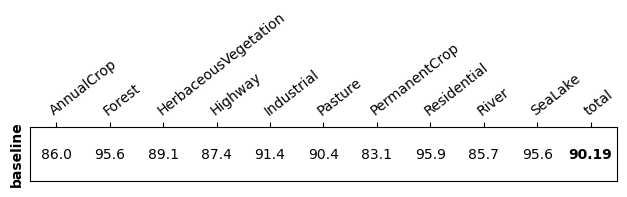}
    \caption{Heatmaps for all classes from the offline attack in Sec. 4.1 for ResNet-50 architecture. Ordered from left to right: CIFAR-10, Fashion-MNIST, and EuroSAT datasets. Ordered from top to bottom: accuracy change of the attacked class, the confounding class, the average of the eight non-attacked/non-confounding classes, and total accuracy of the classifier. Baseline accuracies of the model prior to the poisonous attack are shown below.}
    \label{fig:heatmaps50}
\end{figure*}

\begin{figure*}[t]
    \centering
    \includegraphics[width=0.30\linewidth]{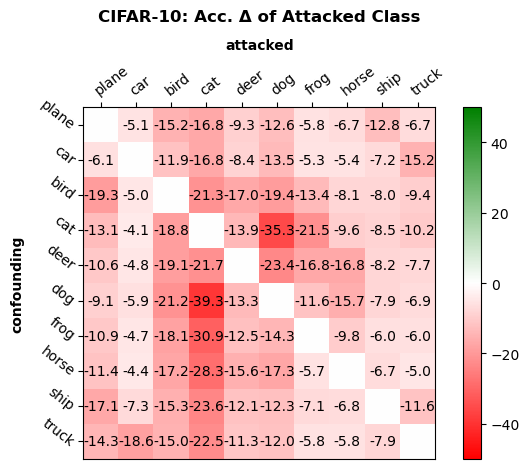}
    \includegraphics[width=0.30\linewidth]{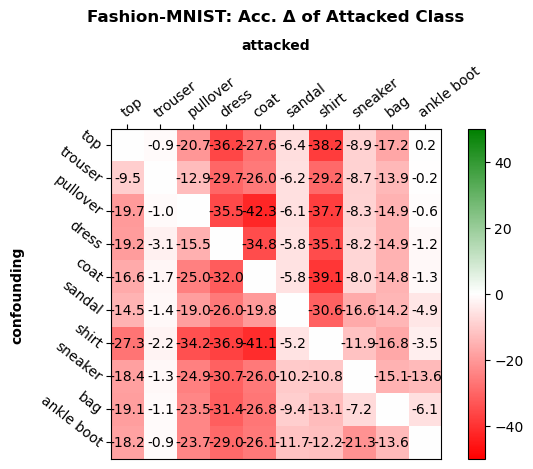}
    \includegraphics[width=0.30\linewidth]{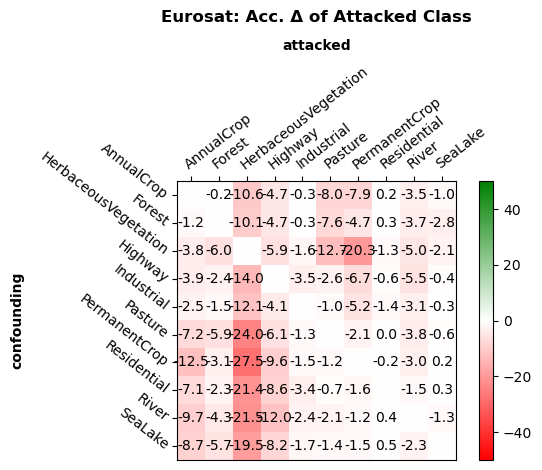}\\
    \includegraphics[width=0.30\linewidth]{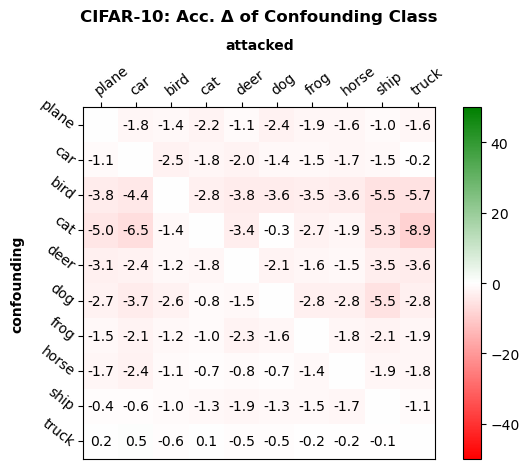}
    \includegraphics[width=0.30\linewidth]{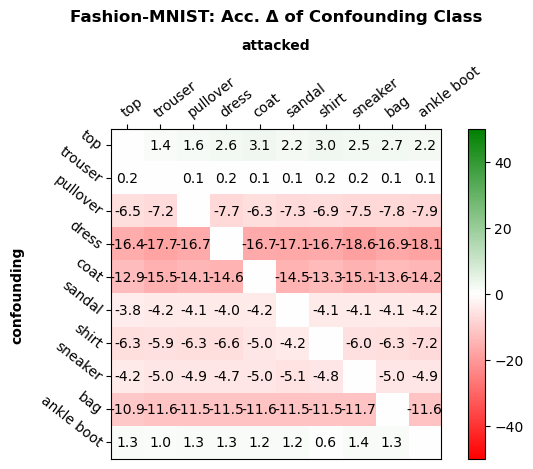}
    \includegraphics[width=0.30\linewidth]{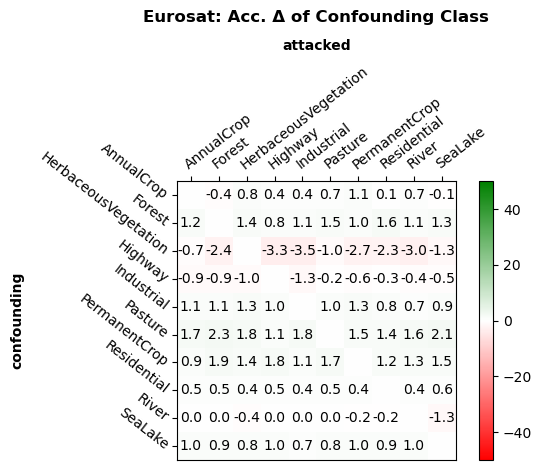}\\
    \includegraphics[width=0.30\linewidth]{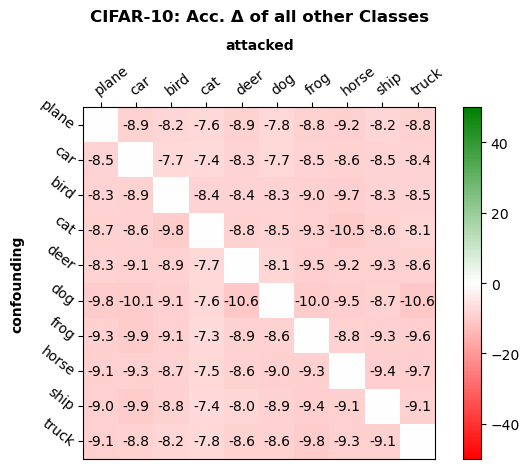}
    \includegraphics[width=0.30\linewidth]{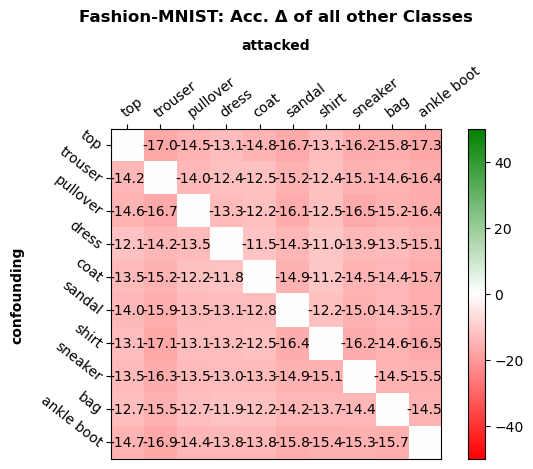}
    \includegraphics[width=0.30\linewidth]{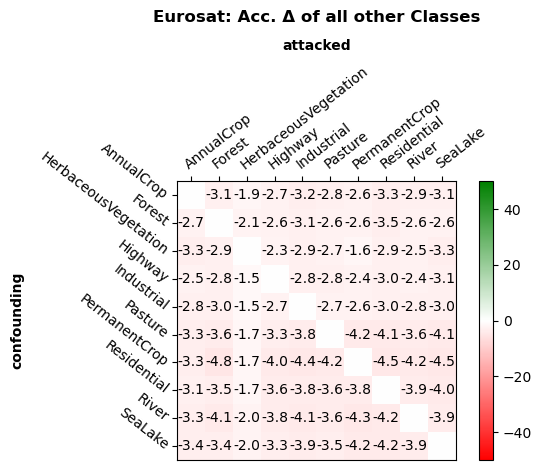}\\
    \includegraphics[width=0.30\linewidth]{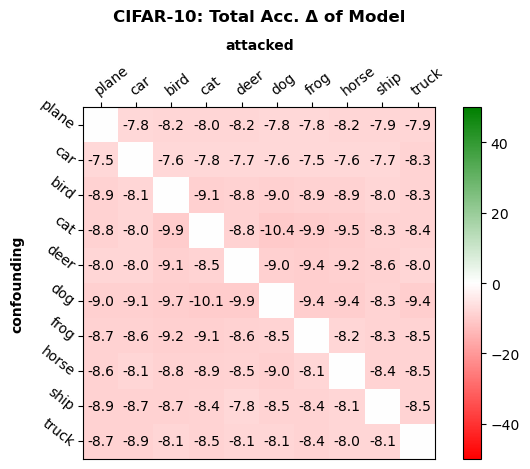}
    \includegraphics[width=0.30\linewidth]{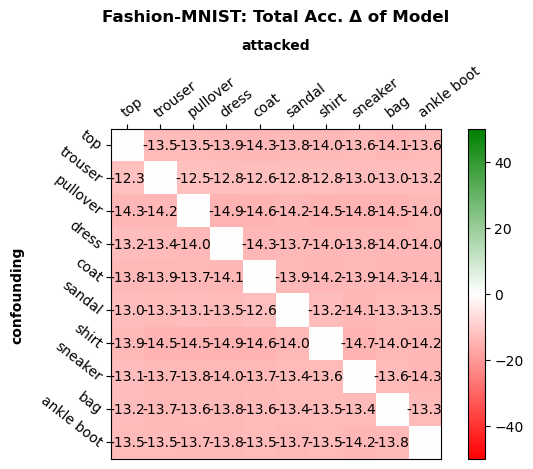}
    \includegraphics[width=0.30\linewidth]{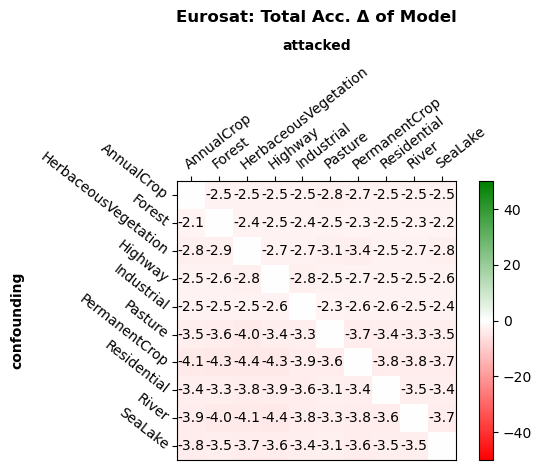}\\
    \includegraphics[width=0.33\linewidth]{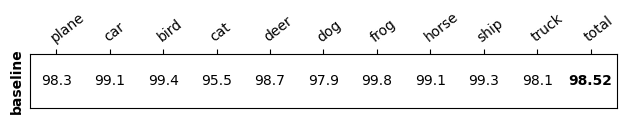}
    \includegraphics[width=0.33\linewidth]{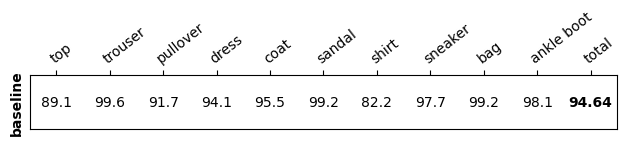}
    \includegraphics[width=0.33\linewidth]{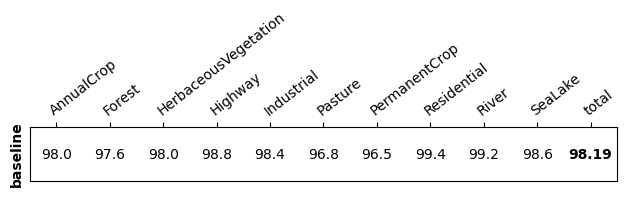}
    \caption{Heatmaps for all classes from the offline attack in Sec. 4.1 for the Vision Transformer (ViT). Ordered from left to right: CIFAR-10, Fashion-MNIST, and EuroSAT datasets. Ordered from top to bottom: accuracy change of the attacked class, the confounding class, the average of the eight non-attacked/non-confounding classes, and total accuracy of the classifier. Baseline accuracies of the model prior to the poisonous attack are shown below.}
    \label{fig:heatmapsvit}
\end{figure*}

In \cref{fig:heatmaps18}, \cref{fig:heatmaps50}, and \cref{fig:heatmapsvit}, we provide the heatmaps for all other classes (attacked, confounding, average of the non-attacked/non-confounding classes, and total accuracy) from the offline mole recruitment attack in Sec. 4.1. We generally observe decreases in accuracy for the attacked class, slight increases in accuracy for the confounding class, no change for the other classes, and negligible change for total accuracy unless in the event of a very strong attack for a particular attacked/confounding combination. However we do observe some general degradation across all classes with ViT, regardless of combination, that occurs during fine-tuning.

\section{Mole Distribution Ablation}

We evaluate the effect of two hyperparameters of our optimized attack in the offline setting:
the correlation $r(\cdot)$ and the threshold $\tilde{\rho}$ at which we stop selecting \texttt{a,c} combinations.
We designate $r_p(\cdot)$ as the correlation derived between $\rho_{a,c}$ and $\mu_{a,c}$ when $\rho_{a,c}$ reflects the $p^{th}$ percentile of all $P(y_a | x_c)$ and $\tilde{\rho}$ as the threshold at which additional combinations should no longer be selected.
We present the results of these attacks in Tab.~\ref{tab:optimized_ablation}.
We observe with decreasing percentiles $p$ in $r_p(\cdot)$, the accuracy penalty generally decreases.
This suggests that the correlation $r(\cdot)$ worsens and selects a less optimal number of $N$ mole samples.
For $r_{95}(\cdot)$ and $\tilde{\rho}{=}0.5$ when using Fashion-MNIST, the correlation appears so poor that the algorithm selects zero samples.
We also observe that while $\tilde{\rho} = 0.1$ appears close to the optimal value, $\tilde{\rho} = 0.01$ can pick up additional combinations that, despite having very low $\rho_{a,c}$, still contribute to a slight accuracy degradation.

\section{Experimental Limitations}
While our experiments explored many different variables to understand the behavior of Mole Recruitment (i.e. \texttt{a,c} class combinations, \# of moles used, softmax probability threshold), we acknowledge there are several variables that should still be explored.
A particularly interesting study would be to explore the effect of batch size with respect to the number of moles used.
We hypothesize the number of gradient steps is critical to the effectiveness of the attack; more gradient steps means greater opportunity to move weights into a position where the model confounds classes.
Thus, spreading the moles over a greater number of batches could yield higher accuracy degradation.

\begin{table}[h]
    \centering
    \small 
    \resizebox{\linewidth}{!}{
    \begin{tabular}{|cc|c|c|c|}
        \hline
        \multicolumn{2}{|c|}{\bf{CIFAR-10}} & [$\rho$'s of selected] & $\tilde{N}$ & $max(\Delta_{total})$\\
        \hline
        \multirow{3}{*}{$\tilde{\rho}=0.01$} & $r_{95}(\cdot)$ & [0.52, 0.02] & 150 & -5.8\\
        & $r_{97}(\cdot)$ & [0.81, 0.6, 0.2, 0.2] & 100 & -8.9\\
        & $r_{99}(\cdot)$ & [0.99, 0.51, 0.21, 0.21] & 200 & -14.0\\
        \hline
        \multirow{3}{*}{$\tilde{\rho}=0.1$} & $r_{95}(\cdot)$ & [0.52] & 200 & -4.1\\
        & $r_{97}(\cdot)$ & [0.82] & 50 & -0.5\\
        & $r_{99}(\cdot)$ & [0.99, 0.51, 0.21, 0.21] & 200 & -14.0\\
        \hline
        \multirow{3}{*}{$\tilde{\rho}=0.5$} & $r_{95}(\cdot)$ & [0.52] & 200 & -4.9\\
        & $r_{97}(\cdot)$ & [0.82] & 50 & -0.5\\
        & $r_{99}(\cdot)$ & [0.99, 0.52] & 200 & -6.9\\
        \hline
        \hline
        \multicolumn{2}{|c|}{\bf{Fashion-MNIST}} & [$\rho$'s of selected] & $\tilde{N}$ & $max(\Delta_{total})$\\
        \hline
        \multirow{3}{*}{$\tilde{\rho}=0.01$} & $r_{95}(\cdot)$ & [0.57, 0.43, 0.24] & 150 & -3.5\\
        & $r_{97}(\cdot)$ & [0.76, 0.45, 0.18] & 150 & -5.1\\
        & $r_{99}(\cdot)$ & [0.93, 0.80, 0.71, 0.02] & 150 & -5.2\\
        \hline
        \multirow{3}{*}{$\tilde{\rho}=0.1$} & $r_{95}(\cdot)$ & [0.57, 0.43, 0.23] & 150 & -3.4\\
        & $r_{97}(\cdot)$ & [0.76, 0.45, 0.18] & 150 & -4.8\\
        & $r_{99}(\cdot)$ & [0.93, 0.80, 0.71] & 200 & -4.8\\
        \hline
        \multirow{3}{*}{$\tilde{\rho}=0.5$} & $r_{95}(\cdot)$ & [0.57] & 0 & 0.0\\
        & $r_{97}(\cdot)$ & [0.76] & 100 & -2.3\\
        & $r_{99}(\cdot)$ & [0.93, 0.80, 0.71] & 200 & -4.8\\
        \hline
    \end{tabular}
    }
    \caption{Total accuracy degradation of an optimized poisonous attack, w.r.t. $f_x(\cdot)$ and $\tilde{\rho}$ hyperparameters, mole set size $\tilde{N}$ and $\rho_{a,c}$ of the selected \texttt{a,c} class combinations.\label{tab:optimized_ablation}}
\end{table}

It would also be interesting to explore how training prior to the attack (i.e. number of epochs, baseline accuracy), could affect the performance of Mole Recruitment.
Perhaps finding a stronger local optima prior to the attack could provide a stronger degree robustness.
Or perhaps alternatively, the poisoning could be just as effective, revealing just how susceptible the weights of machine learning models are to perturbations from naturally occurring data.
Lastly, it would interesting to explore if there other means of identifying poisonous, unaltered data samples, beyond leveraging softmax probabilities and the attacked/confounding framework established in this work.
Such findings could expose even greater vulnerabilities of continual learning models.

\end{document}